\documentclass[11pt]{article}

\usepackage[]{acl}
\usepackage{makecell} 
\usepackage{times}
\usepackage{latexsym}
\usepackage{amsmath}
\usepackage[T1]{fontenc}
\usepackage{amssymb}

\usepackage[utf8]{inputenc}

\usepackage{microtype}

\usepackage{inconsolata}

\usepackage{graphicx}

\usepackage[utf8]{inputenc}
\DeclareUnicodeCharacter{202A}{}
\title{Robust Audio-Text Retrieval via Cross-Modal Attention and Hybrid Loss }
 
 \author{
\textbf{Meizhu Liu \quad Matthew Rowe \quad Amit Agarwal \quad Michael Avendi \quad Yassi Abbasi} \\
\textbf{Paul Li \quad Hitesh Laxmichand Patel \quad Kyu J. Han \quad Tao Sheng \quad Sujith Ravi \quad Dan Roth}
}
\begin{document}
\maketitle

\begin{abstract} 
Audio-text retrieval enables semantic alignment between audio content and natural language queries, supporting applications in multimedia search, accessibility, and surveillance. However, current state-of-the-art approaches struggle with long, noisy, and weakly labeled audio due to their reliance on contrastive learning and large-batch training. We propose a novel multimodal retrieval framework that refines audio and text embeddings using a cross-modal embedding refinement module combining transformer-based projection, linear mapping, and bidirectional attention. To further improve robustness, we introduce a hybrid loss function blending cosine similarity, $\mathcal{L}_{1}$, and contrastive objectives, enabling stable training even under small-batch constraints. Our approach efficiently handles long-form and noisy audio  (SNR 5 to 15) via silence-aware chunking and attention-based pooling. Experiments on benchmark datasets demonstrate improvements over prior methods.
\end{abstract}

\section{Introduction}
Audio data is rapidly growing across domains such as journalism, entertainment, accessibility, and surveillance. Retrieving text from audio and audio from text remains a challenging problem despite its practical importance, for example, detecting glass breaking in a bank or retrieving music clips based on text queries.

These tasks are difficult due to the unstructured, temporally extended, and often weakly labeled nature of audio, which differs fundamentally from images or text. Moreover, real-world recordings frequently contain multiple overlapping events, while accompanying captions typically describe only a subset of them, creating a significant supervision mismatch. For instance, a 30-second recording may include rainfall, background chatter, and animal sounds, yet the caption might mention only “rainfall.” Such discrepancies limit the effectiveness of supervised audio–text retrieval models and make training more challenging.

Some methods use retrieval-augmented audio captioning \cite{Choi25, Ghosh23}, which improves performance but increases training and inference costs. Retrieval from large knowledge bases is expensive. Performance also depends on the quality of retrieved audio–text pairs \cite{petroni2020context, wu2024irrelevant}. Noisy or irrelevant examples can mislead generation, degrade captions \cite{ma2023query, gao2023precise}, and reduce retrieval accuracy.

Recent approaches such as CLAP \cite{clap2023}, ImageBind \cite{imagebind2023}, and Wav2CLIP \cite{wu2022wav2clip} rely on contrastive learning. While effective, they have clear limitations:
\begin{enumerate}
\item \textbf{Annotation sparsity and false negatives:} Weak supervision causes semantically related pairs to be treated as negatives, biasing gradients and harming alignment.
\item \textbf{Limited robustness:} Performance drops on noisy or multi-event audio, where global alignment misses fine-grained semantics.
\item \textbf{Large-batch dependence:} Contrastive losses rely on many in-batch negatives. Small batches yield biased and high-variance gradients \cite{Vaessen24}. Large batches increase memory and compute cost, especially for long audio.
\end{enumerate}

We propose a robust audio–text retrieval framework. It uses cross-modal attention \cite{lei2022loopitr, lei2024mcad, chen2024make, wang2022distilleddualencoder, wang2023conaclip} and a hybrid loss \cite{liu2013, liu2016}. Our contributions are:
\begin{itemize}
\item A \textbf{cross-modal embedding refinement module} with transformer projections and cross-attention for deep semantic alignment.
\item A \textbf{hybrid loss} combining cosine similarity, $\mathcal{L}_{1}$, and contrastive objectives to reduce large-batch dependence.
\item An \textbf{effective long-audio strategy} using silence-aware chunking and attention pooling for noisy, multi-event recordings.
\end{itemize}

Unlike prior CLAP-style methods that rely solely on global contrastive alignment, our approach introduces a training-only cross-modal refinement stage that enables fine-grained semantic interaction without sacrificing dual-encoder efficiency at inference. In addition, we show that combining metric-based alignment losses with contrastive objectives substantially reduces sensitivity to batch size and label noise, an issue largely overlooked in existing audio–text retrieval literature.

\section{Proposed Method}

We present an audio-text retrieval framework. It addresses key limitations (annotation sparsity and false negatives, and large-batch-size–dependent optimization) of existing methods, especially for long, noisy, or weakly labeled audio under small-batch training. Our system uses a multi-stage multimodal architecture. It combines cross-modal embedding refinement with a hybrid loss, improving robustness and semantic alignment.

\subsection{Model Overview}  

Our framework consists of three main components:
\begin{enumerate}
    \item \textbf{Multimodal Encoders:} An audio encoder (e.g. HTSAT-tiny \cite{chen2022htsat}, Whisper \cite{radford2023robust}) and a text encoder (e.g., LLaMA-3B \cite{dubey2024llama3}, BERT \cite{devlin2019bert}, RoBERTa \cite{liu2019roberta}) to extract embeddings from raw audios and texts. These encoders can be replaced with other pretrained backbones to leverage large-scale pretraining.
    \item \textbf{Cross-Modal Embedding Refinement Module:} A three-stage projection module with transformer-based projection, linear transformation, and cross-attention. This approach enables deep, context-aware alignment between audio and text.
    \item \textbf{Hybrid Loss Function:} A weighted combination of cosine similarity, $\mathcal{L}_{1}$, and contrastive loss to improve alignment.
\end{enumerate}

\begin{figure}[t]
  \centering
  \includegraphics[width=\columnwidth]{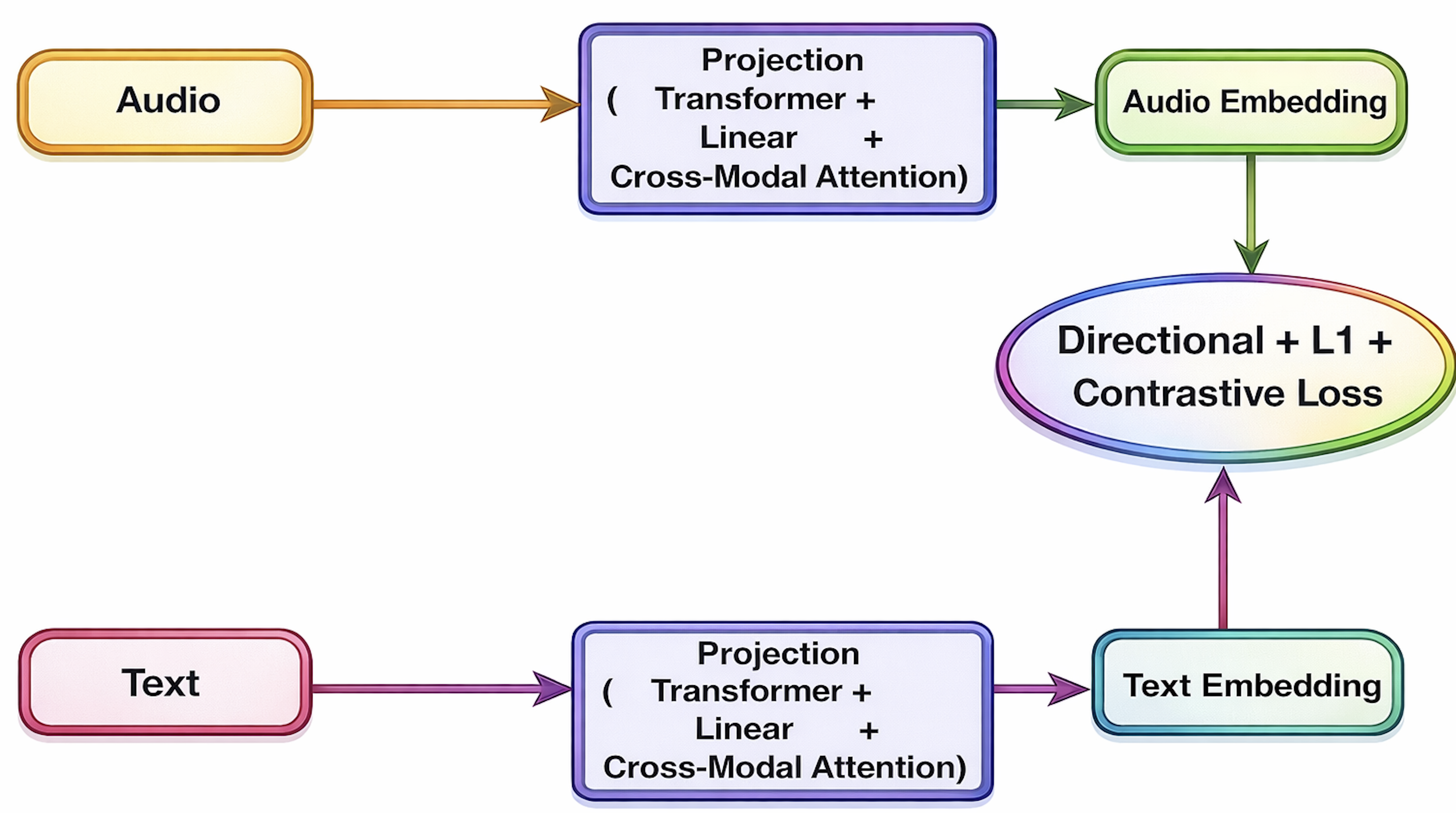}
  \caption{Overview of the proposed audio-text retrieval framework. 
  The projection module refines embeddings through three sequential components: transformer-based projection, linear transformation, and cross-modal attention.
  }
  \label{fig:architecture}
\end{figure}

Figure~\ref{fig:architecture} shows the model architecture. In training, only the projection layers are updated. The audio and text encoders can be frozen or selectively fine-tuned depending on the experiment. Unlike cross-encoder retrieval models, which jointly encode audio–text pairs at inference time and incur quadratic cost, our method uses cross-modal attention only during training. This allows us to retain the scalability and pre-computability advantages of dual encoders.

\subsection{Projection Layers}

Rather than relying on simple linear projections, as used in  LAION-AI CLAP \cite{laion2023}-we
propose a composite alignment module designed to refine and align the embeddings from each
modality in a deeper and more context-aware manner. 
The module has three  sequential components:
 
\textbf{Transformer-based Projection:} 

    Let $\mathbf{X} \in \mathbb{R}^{n \times d_{\text{model}}}$ denote the input embeddings for a given modality, 
    where $n$ is the sequence length and $d_{\text{model}}$ is the embedding dimension. 
    The transformer encoder applies multi-head self-attention \cite{vaswani2017attention} and a feedforward network:
    \begin{align}
        \mathbf{H} &= \text{MHA}(\mathbf{X}) + \mathbf{X}, \\
        \mathbf{Z} &= \text{FFN}(\mathbf{H}) + \mathbf{H},
    \end{align}
    
    
    where MHA denotes multi-head attention with $8$ heads, and FFN is a two-layer feedforward network 
    with hidden dimension $4 \times d_{\text{model}}$ ($d_{\text{model}}$ is the dimension of the embedding model) and GELU activation. 
    Dropout with rate $0.1$ is applied after each sublayer.

 \textbf{Linear Transformation:} 
 
    The transformer-refined embeddings $\mathbf{Z}$ are projected into a shared embedding space 
    of dimension $d_{\text{shared}}$ through a linear transformation:
    \begin{equation}
        \mathbf{E} = \mathbf{Z}\mathbf{W}_{\text{proj}} + \mathbf{b}_{\text{proj}},
    \end{equation}
    where $\mathbf{W}_{\text{proj}} \in \mathbb{R}^{d_{\text{model}} \times d_{\text{shared}}}$ ($d_{\text{shared}}$ is the dimension of the shared embedding space)
    and $\mathbf{b}_{\text{proj}} \in \mathbb{R}^{d_{\text{shared}}}$.

\textbf{Cross-Modal Attention:} 
    Given two modalities, audio embeddings $\mathbf{E}^{a}$ and text embeddings $\mathbf{E}^{t}$, 
    we employ cross-attention such that each modality attends to the other. 
    For example, audio attending to text is computed as:
    

     \begin{align*}
        \mathbf{Q}^{a} &= \mathbf{E}^{a} \mathbf{W}_Q^{a}, \\
        \mathbf{K}^{t} &= \mathbf{E}^{t} \mathbf{W}_K^{t}, \\
        \mathbf{V}^{t} &= \mathbf{E}^{t} \mathbf{W}_V^{t}, \\
        \text{Attn}(\mathbf{E}^{a}, \mathbf{E}^{t}) &=\operatorname{softmax}\!\left(\frac{\mathbf{Q}^{a} (\mathbf{K}^{t})^\top}{\sqrt{d_k}}\right) \mathbf{V}^{t}, \\
    \end{align*}
    where $d_k$ is the key dimension, and $\mathbf{W}$s are learnable projection matrices.
    A symmetric formulation is applied for text attending to audio. 
    This bidirectional attention enables the model to extract semantically aligned features across modalities.
    
    Note although cross-modal attention is used during training to encourage fine-grained alignment between audio and text representations, it is not used at inference time. For retrieval, embeddings for each modality are computed independently using the transformer-based projection and linear mapping only. This design preserves the efficiency and scalability of dual-encoder retrieval while benefiting from cross-modal interaction during training. 

\subsection{Hybrid Loss Function} 

To address the limitations of contrastive learning \cite{clap2023, laion2023} in small batches and noisy labeling, we propose a
custom loss function which is a weighted combination of the following three types of losses:

\begin{itemize}
    \item Directional loss: this is the cosine similarity loss, and it encourages audio and text embeddings to align along
    semantically meaningful directions.
    \item $\mathcal{L}_{1}$ loss: promotes fine-grained consistency between refined embeddings, with robustness to noise and outliers.
    \item Contrastive loss: maintains discriminative alignment between matched and unmatched pairs but
    is complemented by the above losses to reduce dependency on batch size.
\end{itemize} 

The hybrid loss function is written as
\begin{equation}
\mathcal{L}_{\text{hybrid}} = 
\lambda_{1} \, \mathcal{L}_{\text{dir}} \;+\; 
\lambda_{2} \, \mathcal{L}_{1} \;+\; 
\lambda_{3} \, \mathcal{L}_{\text{con}} ,
\label{eq:loss}
\end{equation}
where $\lambda_{1}, \lambda_{2}, \lambda_{3} \geq 0$ and $\lambda_{1}+\lambda_{2} + \lambda_{3}=1$ are weighting hyperparameters that control the contribution of each loss component. These can be
optimized based on the available batch size and dataset. We will cover this in detail in Section~\ref{sec:experiments}. 

Contrastive loss alone optimizes relative similarity but provides no constraint on absolute embedding alignment, making optimization unstable under small batch sizes and noisy negatives. The cosine term stabilizes directional alignment, while the L1 loss enforces local consistency between matched pairs, acting as a regularizer that reduces gradient variance. Together, these objectives balance global discrimination with local alignment.

\subsection{Handling Long Audios}
 \label{sec:long_audio}
 
Our model handles longer audio input via  silence removal, chunking, followed by 
attention-based pooling to capture the most relevant audio segments. 
This design makes the model scalable and effective for real-world audio data.

\paragraph{Chunking:} 
For long audio inputs ($>10$ seconds), we first remove extended silences 
(greater than $1$ second, which we found to be an effective threshold), 
and then divide the remaining audio into overlapping or fixed-length chunks 
(e.g., $10$-second segments. We used fixed-length in our experiments). 

Formally, let the preprocessed audio signal be $\mathbf{x}$, and 
$\{\mathbf{x}_1, \mathbf{x}_2, \dots, \mathbf{x}_m\}$ denote the resulting 
chunks after segmentation. Each chunk $\mathbf{x}_i$ is independently encoded by the audio encoder:
\begin{equation}
    \mathbf{h}_i = f_{\text{enc}}(\mathbf{x}_i), \quad i = 1, \dots, m,
\end{equation}
where $f_{\text{enc}}$ is the audio encoder and $\mathbf{h}_i$ is the 
chunk-level embedding. This chunked encoding process allows the model 
to handle arbitrarily long audio sequences.

\begin{table*}[h]
\centering
\begin{tabular}{llcccccc}
\hline
\textbf{Model} & \textbf{Dataset} & \textbf{Modality} & \textbf{R@1} & \textbf{R@5} & \textbf{R@10} & \textbf{mAP@10} \\
\hline
Microsoft-CLAP & Clotho    & a2t & 0.232 & 0.475 & 0.576 & 0.154 \\
LAION-CLAP     & Clotho    & a2t & 0.175 & 0.370 & 0.455 & 0.155 \\
\textbf{Proposed (ours)} & Clotho    & a2t & \textbf{0.183} & \textbf{0.482} & \textbf{0.591} & \textbf{0.162} \\

Microsoft-CLAP & AudioCaps & a2t & 0.381 & 0.697 & 0.814 & 0.319 \\
LAION-CLAP     & AudioCaps & a2t & 0.444 & 0.768 & 0.889 & 0.438 \\
\textbf{Proposed (ours)} & AudioCaps & a2t & \textbf{0.451} & \textbf{0.793} & \textbf{0.905} & \textbf{0.486} \\

Microsoft-CLAP & ESC50     & a2t & 0.935 & 0.998 & 1.000 & 0.956 \\
LAION-CLAP     & ESC50     & a2t & 0.915 & 0.995 & 0.997 & 0.947 \\
\textbf{Proposed (ours)} & ESC50     & a2t & \textbf{0.950} & \textbf{0.995} & \textbf{0.998} & \textbf{0.972} \\

Microsoft-CLAP & FSD50K    & a2t & 0.542 & 0.837 & 0.897 & 0.581 \\
LAION-CLAP     & FSD50K    & a2t & 0.655 & 0.874 & 0.913 & 0.659 \\
\textbf{Proposed (ours)} & FSD50K    & a2t & \textbf{0.697} & \textbf{0.889} & \textbf{0.929} & \textbf{0.672} \\
\hline
Microsoft-CLAP & Clotho    & t2a & 0.156 & 0.385 & 0.510 & 0.255 \\
LAION-CLAP     & Clotho    & t2a & 0.146 & 0.349 & 0.447 & 0.231 \\
\textbf{Proposed (ours)} & Clotho    & t2a & \textbf{0.158} & \textbf{0.420} & \textbf{0.543} & \textbf{0.267} \\

Microsoft-CLAP & AudioCaps & t2a & 0.289 & 0.630 & 0.770 & 0.432 \\
LAION-CLAP     & AudioCaps & t2a & 0.341 & 0.697 & 0.827 & 0.490 \\
\textbf{Proposed (ours)} & AudioCaps & t2a & \textbf{0.352} & \textbf{0.715} & \textbf{0.844} & \textbf{0.521} \\
\hline
\end{tabular}
\caption{Audio--text retrieval results across datasets and models. The \textbf{Modality} column indicates audio-to-text (a2t) or text-to-audio (t2a) retrieval. Metrics include Recall@K ($K=1,5,10$) and mAP@10 (higher is better).}
\label{tab:retrieval_results}
\end{table*}

\paragraph{Attention-Based Pooling:}
Instead of averaging, we apply attention-based pooling. Given chunk embeddings ${\mathbf{h}_1, \dots, \mathbf{h}_m}$, the pooled embedding is
\begin{equation}
\mathbf{z} = \sum_{i=1}^{m} \alpha_i \mathbf{h}_i, \\
\alpha_i= \frac{\exp(g(\mathbf{h}_i, \mathbf{q}))}{\sum_{j=1}^{m} \exp(g(\mathbf{h}_j, \mathbf{q}))}
\label{eqn:pooling}
\end{equation}
where $g(\cdot)$ is a relevance scoring function and $\mathbf{q}$ is the attention query.


Attention-based pooling is particularly useful when captions describe only part of the audio. For example, if an audio clip contains rain, sheep, and chatter but the caption mentions only rain, attention focuses on rain. This makes the model robust to noisy, multi-event recordings where uniform pooling fails. Implementation details of the training and evaluation procedures for attention-based pooling are provided in ~\ref{sec:pooling}.

Together, silence-aware chunking and attention-based pooling bound the effective audio sequence length, keeping training-time cross-modal interactions computationally manageable while enabling robust retrieval from long, noisy, and multi-event audio recordings.

\subsection{Computational Complexity}

During training, our framework introduces cross-modal attention to enable fine-grained alignment between audio and text representations. This interaction incurs a computational cost of $\mathcal{O}(n_a n_t)$, where $n_a$ and $n_t$ denote the numbers of audio and text tokens, respectively. As described in Section~\ref{sec:long_audio}, the audio sequence length $n_a$ is bounded through silence-aware chunking and attention-based pooling, which limit the number of chunk-level embeddings regardless of the raw audio duration, ensuring that the training-time cost remains practical even for long or multi-event recordings.

Crucially, cross-modal attention is applied only during training. At inference time, audio and text inputs are encoded independently, using the same transformer-based projection and linear mapping as the dual-encoder baseline. Consequently, the inference complexity remains $\mathcal{O}(n)$ ($n$ refers to the number of tokens in a single modality, i.e., the sequence length of either the audio or text input that is fed into the encoder at inference time), preserving the efficiency and scalability characteristic of standard dual-encoder retrieval models such as CLAP, while still benefiting from richer cross-modal supervision during training.

\section{Experimental Results}
\label{sec:experiments}

We perform hyperparameter tuning using Optuna \cite{optuna2019}. The tuned hyperparameters include the number of layers and attention heads in the transformer projection, the feedforward dimension, dropout rate, activation function (GELU or ReLU), dimensions of the shared embedding space, number of training epochs, and the weights in the weighted loss function (Eq.~\ref{eq:loss}). More training details can be found in \ref{app:trainingdetails}. For all experiments, we use HTSAT-tiny as the audio encoder and RoBERTa-large as the text encoder, keeping both encoders frozen while training only the projection layer. Additional experiments and ablation studies on the encoders are provided in  \ref{sec:ablation}.


\begin{table}[h]
\centering
\begin{tabular}{|l|c|c|c|}
\hline
\textbf{Model} & \textbf{Dataset}  & \small{\textbf{SNR}} & \small{\textbf{mAP@10}  }\\ 
Microsoft-CLAP & Clotho & 5 &  0.142 \\ 
LAION-CLAP & Clotho &  5 & 0.146 \\ 
Proposed & Clotho  &  5 & 0.161 \\ 
Microsoft-CLAP & Clotho & 10 &  0.134 \\ 
LAION-CLAP & Clotho &  10 & 0.140 \\ 
Proposed & Clotho  &  10 & 0.160 \\ 
Microsoft-CLAP & Clotho & 15 &  0.117 \\ 
LAION-CLAP & Clotho &  15 & 0.125 \\ 
Proposed & Clotho  &  15 & 0.158 \\ 
Microsoft-CLAP & AudioCaps &5 & 0.265  \\ 
LAION-CLAP & AudioCaps &5 &0.402  \\ 
Proposed & AudioCaps  & 5 &0.474 \\ 
Microsoft-CLAP & AudioCaps &10 & 0.242  \\ 
LAION-CLAP & AudioCaps &10 &0.347  \\ 
Proposed & AudioCaps  & 10 &0.468 \\ 
Microsoft-CLAP & AudioCaps &15 & 0.203  \\ 
LAION-CLAP & AudioCaps &15 &0.317  \\ 
Proposed & AudioCaps  & 15 &0.466 \\ 
Microsoft-CLAP & ESC50 &  5 & 0.951 \\ 
LAION-CLAP & ESC50 &   5 & 0.942 \\ 
Proposed & ESC50  &  5 & 0.970   \\ 
Microsoft-CLAP & ESC50 & 10 & 0.904 \\ 
LAION-CLAP & ESC50 &10 & 0.872 \\ 
Proposed & ESC50  &10 & 0.936   \\ 
Microsoft-CLAP & ESC50 & 15 & 0.812 \\ 
LAION-CLAP & ESC50 &15 & 0.797 \\ 
Proposed & ESC50  &15 & 0.932   \\ 
Microsoft-CLAP & FSD50K &  5 &  0.564\\ 
LAION-CLAP & FSD50K &5 & 0.625  \\ 
Proposed & FSD50K  &5 & 0.661   \\ 
Microsoft-CLAP & FSD50K &  10 &0.402 \\ 
LAION-CLAP & FSD50K &10 & 0.479  \\ 
Proposed & FSD50K  &10 &0.593  \\ 
Microsoft-CLAP & FSD50K &  15 &0.378 \\ 
LAION-CLAP & FSD50K &15 & 0.401  \\ 
Proposed & FSD50K  &15 &0.588  \\ 
\hline
\end{tabular}
\caption{Audio to text retrieval on  datasets with different levels (SNR ranges from 5 to 15) of noise.}
\label{tab:a2tresultsnoisy}
\end{table}

\begin{table}[h]
\centering
\begin{tabular}{|l|c|c|c|c|}
\hline
\textbf{Model} & \textbf{Dataset}  &\small{\textbf{SNR}} & \small{\textbf{mAP@10}}  \\ 
Microsoft-CLAP & Clotho & 5& 0.252  \\ 
LAION-CLAP & Clotho &  5& 0.227\\ 
Proposed & Clotho  &  5& 0.256 \\ 
Microsoft-CLAP & Clotho & 10& 0.223  \\ 
LAION-CLAP & Clotho &  10& 0.217\\ 
Proposed & Clotho  & 10& 0.255 \\ 
Microsoft-CLAP & Clotho & 15& 0.207  \\ 
LAION-CLAP & Clotho &  15& 0.201\\ 
Proposed & Clotho  & 15& 0.252 \\ 
Microsoft-CLAP & AudioCaps &5&  0.421\\ 
LAION-CLAP & AudioCaps &5& 0.475 \\ 
Proposed & AudioCaps  &5& 0.520 \\ 
Microsoft-CLAP & AudioCaps &10& 0.405 \\ 
LAION-CLAP & AudioCaps &10&0.437    \\ 
Proposed & AudioCaps  &10& 0.518 \\ 
Microsoft-CLAP & AudioCaps &15& 0.386 \\ 
LAION-CLAP & AudioCaps &15&0.408    \\ 
Proposed & AudioCaps  &15& 0.510 \\ 
\hline
\end{tabular}
\caption{Text to audio retrieval on noisy datasets with different levels of noise (SNR ranges from 5 to 15).}
\label{res:t2aresultsnoisy}
\end{table}

\begin{table*}[h]
\centering
\small
\renewcommand{\arraystretch}{1.3}
\setlength{\tabcolsep}{4pt}
\begin{tabular}{|p{0.26\textwidth}|p{0.14\textwidth}|p{0.1\textwidth}|p{0.36\textwidth}|}
\hline
\textbf{Audio Clip Description} & \textbf{Text Query} & \textbf{Retrieval} & \textbf{Comments} \\
\hline
Rain, birds chirping, dog barking & ``rain sounds'' & Good & Attention focuses on the relevant rain segment despite background events \\
Two people talking, laughter, footsteps & ``people talking and laughing'' & Good & Correctly retrieves despite extra footsteps in the audio \\
Engine noise, wind, distant horn & ``car engine'' & Bad & Retrieves a clip with wind noise; subtle engine sound missed, showing limitations for low-SNR events \\
Piano playing with applause in background & ``piano music'' & Good & Correct retrieval; cross-modal attention helps focus on piano despite applause \\
Child laughing, hands clapping, toy squeaking, dog barking & ``dog barking'' & Bad & Background noise and multiple events cause partial mismatch; demonstrates current attention pooling limitations \\
\hline
\end{tabular}
\caption{Example audio–text retrieval results. ``Good'' indicates successful retrieval where the model correctly identifies the queried event, while ``Bad'' indicates retrieval failures.}
\label{tab:examples}
\end{table*}

\subsection{Data and Results}

We train and evaluate our model on publicly available datasets using their standard training and test splits, including FSD50K \cite{fonseca2022}, ESC-50 \cite{esc50}, Clotho \cite{clotho2020}, and AudioCaps \cite{audiocaps2019}. We compare against Microsoft-CLAP \cite{clap2023} and LAION-CLAP \cite{laion2023}, which provide publicly available checkpoints. For each baseline, we evaluate all released checkpoints as well as versions fine-tuned on the corresponding training data (using the code provided at their githubs), and report the best-performing results.

Audio-to-text retrieval was evaluated on the test splits of all datasets. Text-to-audio retrieval was evaluated on Clotho and AudioCaps. Results  are reported in Table~\ref{tab:retrieval_results}. Across both retrieval directions and all test sets, our model generally achieves higher mAP and Recall@K than the baselines.  
Since all models are evaluated on the same test audios, we further assess statistical significance using a paired Wilcoxon signed-rank test \cite{wilcoxon1945individual} on per-audio AP@10 scores across all datasets, comparing our model with the second-best baseline. The $p$ value indicates that the observed improvements are statistically significant for both audio-to-text retrieval ($p=0.011$) and text-to-audio retrieval ($p=0.009$).

\subsection{Robust Evaluation}

We further evaluate robustness by adding randomly sampled noise from the MUSAN corpus \cite{snyder2015musan} at varying signal-to-noise ratios (SNRs) and evaluate all models on the resulting noisy datasets. The results for audio to text and text to audio retrieval are shown in Table \ref{tab:a2tresultsnoisy} and \ref{res:t2aresultsnoisy} respectively. 
While all methods degrade under noise, our approach shows substantially smaller performance drops, suggesting that cross-modal refinement and attention-based pooling mitigate the effect of irrelevant or corrupted acoustic segments.




\subsection{Ablation Studies}


We conduct ablation studies by systematically varying key components of our model, including audio encoders, text encoders, projection layers, batch sizes, loss functions, and weight configurations for the combined loss. As shown in Table~\ref{tab:ablation1} in \ref{sec:ablation}, transformer-based projection layers paired with the combined loss consistently yields the strongest performance, while the model remains robust across batch sizes. Both audio-to-text and text-to-audio retrieval tasks benefit from this configuration, highlighting the advantages of more expressive projection mechanisms and multi-objective training.

\subsection{Baseline: Caption Based Retrieval}

We establish an additional baseline by generating audio captions followed by text based search. To avoid model training or fine-tuning, we use large pretrained models to produce the audio descriptions. Retrieval is then performed via search (lexical search, semantic based search and BM25 \cite{bm25}) on these captions.  
More details and results can be found in Appendix \ref{sec:semanc}. 
The results indicate that generating captions first and performing retrieval based on them is not very effective.

\subsection{Qualitative Analysis}

To further illustrate the effectiveness of our model, we present some retrieval examples on multi-event audio clips from the AudioCaps and Clotho datasets. Each audio clip contains multiple overlapping events, while captions often describe only a subset. We show successful (\textit{Good}) and failure (\textit{Bad}) retrievals to highlight strengths and limitations.
Our model generally succeeds in retrieving clips where the query matches a \emph{dominant audio event}. Failures often occur when the target sound is weak or masked by louder events, highlighting potential avenues for improving attention-based pooling or chunk weighting. These examples support our claim that cross-modal attention improves fine-grained semantic alignment, particularly for multi-event and noisy recordings.

\section{Conclusions} 

We present a robust framework for audio–text retrieval that integrates multimodal encoders, cross-modal embedding refinement, and a hybrid loss function. This design enables strong semantic alignment across modalities while maintaining robustness to noise and effectiveness under small-batch training regimes. Experimental results demonstrate the effectiveness of cross-modal attention mechanisms and hybrid loss formulations for scalable, high-performance audio–text retrieval. Future work will explore larger multimodal architectures, more complex acoustic environments, and a broader range of audio domains.

\section{Limitations} 

While our framework demonstrates strong performance, several limitations remain:

\begin{itemize}
    \item \textbf{Dependence on Pretrained Encoders:} Our model relies on high-quality pretrained audio and text encoders. Performance may degrade if smaller or less capable encoders are used.
    \item \textbf{Noisy and Complex Audio:} Although our model is robust to moderate levels of additive noise sampled from MUSAN \cite{snyder2015musan}, further evaluation is needed under other real-world noise conditions, such as environmental noise and mixtures of speech and background noise. Moreover, retrieval performance may degrade in extremely noisy settings or when multiple audio events heavily overlap.
    \item \textbf{Silence-based Segmentation} is a coarse heuristic and may fail for continuous background or polyphonic scenes. We need to explore more adaptive segmentation for long-audios.
\end{itemize}

Addressing these limitations is an important direction for future work to improve generalization, efficiency, and robustness across diverse audio environments.

\section{Ethical considerations} 

Our work raises several ethical considerations that should be acknowledged:

\begin{itemize}
    \item \textbf{Privacy Concerns:} Audio data can contain sensitive personal information. Care must be taken to ensure proper consent, anonymization, and compliance with privacy regulations when collecting or using such data.
    \item \textbf{Bias and Fairness:} Pretrained encoders may inherit biases from their training data, potentially affecting retrieval performance across different languages, accents, or demographic groups.
    \item \textbf{Misuse Risks:} Audio-text retrieval systems could be misused for surveillance or other privacy-invasive applications. Responsible deployment and clear usage policies are critical.
    \item \textbf{Environmental Impact:} Training large multimodal models consumes significant computational resources, contributing to energy usage and carbon footprint.
\end{itemize}

Researchers and practitioners should consider these aspects when developing, deploying, or extending audio-text retrieval systems to ensure ethical and responsible use.

\bibliography{bib}

\appendix

\section{Appendix}

\subsection{Attention Based Pooling: Training and Testing Details}
\label{sec:pooling}

During training, $\mathbf{q}$ in Eqn. \ref{eqn:pooling} is the paired text embedding, which encourages the model to focus on semantically relevant audio segments. During inference, $\mathbf{q}$ is replaced by a learnable, query-independent vector $\mathbf{q}_{\text{pool}}$. This enables audio embeddings to be precomputed efficiently. The vector $\mathbf{q}_{\text{pool}}$ is optimized jointly with the rest of the model via backpropagation. Gradients from the retrieval objective update $\mathbf{q}_{\text{pool}}$, allowing it to capture dataset-level priors. 
To mitigate the mismatch between training and inference, the text-conditioned query was randomly replaced with $\mathbf{q}_{\text{pool}}$  during training with a probability of (e.g. 10\%. which can be optimized via Optuna and we found 10\% gave the best results on the validation splits). This strategy encouraged more robust, query-independent pooling behavior.

\subsection{Ablation Studies on Model Components}
\label{sec:ablation}

We performed ablation studies to evaluate the contributions of each major component in our model, including the audio encoder, text encoder, projection layer (linear MLP vs. transformer-based), and loss function (standard contrastive loss vs. our proposed combined loss). As shown in Table \ref{tab:ablation1} (results reported for the Clotho dataset; ablations for additional datasets will be released on our GitHub), the transformer-based projection layer together with the proposed loss consistently yields the strongest performance. This setup improves both audio-to-text and text-to-audio retrieval, highlighting the advantages of more expressive projection architectures and multi-objective optimization.

We further examined the weighting strategy used in the combined loss function (Eq.~\ref{eq:loss}). Table~\ref{tab:ablationweights} shows that using appropriately tuned weights leads to a substantial performance improvement. Additionally, we conducted an ablation study on batch size, with results presented in Table~\ref{tab:batchsize}, demonstrating the model’s robustness to smaller batch sizes.


\begin{table*}[h]
\centering
\begin{tabular}{lccccccccc}
\hline
\textbf{Audio encoder} & \textbf{Text encoder} & \textbf{Projection}  & \textbf{Loss function}   & \textbf{\makecell{a2t \\ mAP@10}}  & \textbf{\makecell{t2a\\ mAP@10}} \\ 
\hline
HTSAT-tiny & Roberta-base & linear & contrastive &  0.134 & 0.182 \\
HTSAT-tiny & Roberta-base & transformer & contrastive &  0.151 & 0.238\\
HTSAT-tiny & Roberta-base & transformer & combined &  0.160 & 0.255 \\

HTSAT-tiny & Roberta-large & linear & contrastive &  0.147 & 0.243 \\
HTSAT-tiny & Roberta-large & transformer & contrastive &  0.155&0.244\\
HTSAT-tiny & Roberta-large & transformer &combined &\textbf{0.162} & 0.267 \\

HTSAT-base & Roberta-base & linear & contrastive &  0.142 & 0.216 \\
HTSAT-base & Roberta-base & transformer & contrastive & 0.153&0.232\\
HTSAT-base & Roberta-base & transformer & combined & 0.161 & 0.254 \\

HTSAT-base & Roberta-large & linear & contrastive &  0.139 & 0.231 \\
HTSAT-base & Roberta-large & transformer & contrastive &0.151&0.245\\
HTSAT-base & Roberta-large & transformer & combined &  0.159 & \textbf{0.259}\\

HTSAT-tiny & Llama 3.2 1B & linear & contrastive &0.143 & 0.199 \\
HTSAT-tiny & Llama 3.2 1B & transformer & contrastive &  0.151 & 0.232\\
HTSAT-tiny & Llama 3.2 1B & transformer & combined & 0.160 & 0.256 \\

HTSAT-tiny & Llama 3.2 3B & linear & contrastive & 0.137 & 0.204 \\
HTSAT-tiny & Llama 3.2 3B & transformer & contrastive & 0.152 & 0.241 \\
HTSAT-tiny & Llama 3.2 3B & transformer & combined & 0.161 & 0.258 \\

HTSAT-base & Llama 3.2 1B & linear & contrastive & 0.139 & 0.211 \\
HTSAT-base & Llama 3.2 1B & transformer & contrastive & 0.142 & 0.236 \\
HTSAT-base & Llama 3.2 1B & transformer & combined & 0.159 & 0.254 \\

HTSAT-base & Llama 3.2 3B & linear & contrastive & 0.139 & 0.242 \\
HTSAT-base & Llama 3.2 3B & transformer & contrastive & 0.148 & 0.253 \\
HTSAT-base & Llama 3.2 3B & transformer & combined & 0.160 & 0.256 \\

Whisper & Roberta-base  & linear & contrastive & 0.149 & 0.238 \\
Whisper & Roberta-base  & transformer & contrastive & 0.152 & 0.249 \\
Whisper & Roberta-base  & transformer & combined & 0.159 & 0.253 \\

Whisper & Roberta-large & linear & contrastive & 0.151 & 0.242 \\
Whisper & Roberta-large & transformer & contrastive & 0.153 & 0.247 \\
Whisper & Roberta-large & transformer & combined & 0.157 & 0.252 \\

Whisper & Llama 3.2 3B & linear & contrastive & 0.141 & 0.240 \\
Whisper & Llama 3.2 3B & transformer & contrastive & 0.148 & 0.247 \\
Whisper & Llama 3.2 3B & transformer & combined & 0.156 & 0.252 \\

Whisper & Llama 3.2 1B & linear & contrastive & 0.136 & 0.243 \\
Whisper & Llama 3.2 1B & transformer & contrastive & 0.142 & 0.249 \\
Whisper & Llama 3.2 1B & transformer & combined & 0.153 & 0.260 \\
\hline 
\end{tabular}
\caption{Ablation studies with different audio and text encoders, different projection types, and different loss functions on the Clotho dataset.}
\label{tab:ablation1}
\end{table*}

\begin{table}[h]
\centering
\begin{tabular}{lcccc}
\hline
\textbf{$\lambda_{1}$} & \textbf{$\lambda_{2}$} & {$\lambda_{3}$}  & \textbf{\makecell{a2t \\ mAP@10}}  & \textbf{\makecell{t2a\\ mAP@10}} \\ 
\hline
0 & 0 & 1 & 0.147 & 0.243 \\
0.1 & 0.2 & 0.7 &  0.149 & 0.250 \\
0.2 & 0.3 & 0.5 & 0.158  & 0.256 \\
0.2 & 0.1 & 0.7 & 0.151 & 0.244\\
\textbf{0.3} & \textbf{0.3} & \textbf{0.4} & \textbf{0.162} & \textbf{0.267} \\ 
0.4 & 0.4 & 0.2 & 0.148 & 0.242 \\ 
\hline 
\end{tabular}
\caption{Ablation studies with different weight combinations for the combined loss function, on the Clotho dataset using HSTAT-tiny and Roberta-large encoders. The Optuna optimized weights are \textbf{0.3}, \textbf{0.3} and \textbf{0.4}.}
\label{tab:ablationweights}
\end{table}


\begin{table}[h]
\centering
\begin{tabular}{lcc}
\hline
\textbf{Batch size} & \textbf{\makecell{a2t \\ mAP@10}}  & \textbf{\makecell{t2a\\ mAP@10}} \\ 
\hline
4  & 0.147 & 0.243 \\
8 &  0.149 & 0.250 \\
16 &  0.158  & 0.256 \\
32 & 0.157 & 0.254\\
64 & 0.156& 0.255 \\ 
\hline 
\end{tabular}
\caption{Ablation studies with different batch sizes, on the Clotho dataset using HSTAT-tiny and Roberta-large encoders. }
\label{tab:batchsize}
\end{table}

\subsection{Audio Caption Generation and Retrieval}
\label{sec:semanc}


Several state-of-the-art audio–language models are capable of generating audio captions, including EnCLAP \cite{kim2024enclap},  Pengi \cite{deshmukh2023pengi}, LTU \cite{ltu}, Qwen-Audio \cite{chu2024qwen2}, GAMA \cite{ghosh2024gama}, AudioGPT \cite{huang2024audiogpt}, Salmonn \cite{tang2023salmonn}, Audio Flamingo \cite{ghosh2025audio, Kong24}, and Gemini families (e.g. Gemini Flash v2, Gemini Pro v1.5, Gemini 2.5 Pro). These models differ in how they extract audio features and integrate them into a large language model. Qwen-Audio and Salmonn primarily focus on speech-related tasks, whereas Pengi, LTU, Audio Flamingo, and Gemini 2.5 Pro emphasize non-speech audio understanding. 
For each of the aforementioned model, we generate 20 captions per audio clip with different prompts. The prompts are designed to capture different levels of detail, context, and diversity. 

\subsubsection{Prompts for audio caption generation}
The prompts we used for generating audio captions are listed below:

\begin{enumerate}
    \item \textbf{Simple descriptive prompt:} \\
    \textit{``Listen to the audio and describe it briefly. Include the main sounds and events you hear.''}
    
    \item \textbf{Detailed multievent prompt:} \\
    \textit{``You will hear an audio clip. Provide a detailed caption describing all audible events, such as voices, music, environmental sounds, and other background noises. Be as specific as possible.''}
    
    \item \textbf{Context-aware prompt:} \\
    \textit{``Listen carefully to the audio clip. Describe the scene as if you are explaining it to someone who cannot hear it. Include people, objects, and actions that can be inferred from the sounds.''}
    
    \item \textbf{Multi-caption prompt:} \\
    \textit{``Generate four different captions for the following audio. Each caption should focus on different aspects of the sound, such as background noises, foreground events, or emotional tone.''}
     
    \item \textbf{Summarization prompt:} \\
    \textit{``After listening to the audio, provide a concise summary of what is happening in no more than 100 words.''}

    \item \textbf{Narrative prompt:} \\
    \textit{``Imagine you are writing a short scene based on this audio. Describe it in a vivid, narrative style, capturing both sounds and atmosphere.''}

    \item \textbf{Freestyle prompt:}  \cite{Kong24} \\
    Can you briefly describe what you hear in this audio?

\end{enumerate}

Some representative examples are shown in Table~\ref{tab:captions}. A comparison with the ground-truth captions reveals substantial discrepancies, highlighting the challenges of accurately capturing audio events and their semantic context.

\begin{table*}[ht]
\centering
\small
\renewcommand{\arraystretch}{1.3}
\setlength{\tabcolsep}{4pt}

\begin{tabular}{|p{0.28\textwidth}|p{0.28\textwidth}|p{0.28\textwidth}|}
\hline
\textbf{LTU} & \textbf{Gemini 2.5 Pro} & \textbf{groundtruth} \\
\hline
a man speaking with laughter and birds singing in the background & 
Gentle voices encouraging a bleating newborn animal, likely a lamb or kid, that is attempting to walk. & 
two women talking then laughing as fabric shuffles followed by a goat baaing \\
\hline
someone claps and drops something on concrete & 
The sound of water splashing forcefully, like from a tap or shower & 
A child giggling as something hits the ground and splatters \\
\hline
an engine chugging followed by ringing & 
A low electronic hum and a growing rumble culminate in a sharp, abrupt hiss. & 
footsteps running on dirt followed by compressed air spraying as wind blows in the background \\
\hline
\end{tabular}
\caption{Examples of audio captions generated by \texttt{LTU} and \texttt{Gemini 2.5 Pro}. The column labeled \textbf{groundtruth} contains reference human-written captions. The discrepancies between model outputs and ground truth highlight the challenges of accurate audio caption generation.}
\label{tab:captions}
\end{table*}

\subsubsection{Caption filtering and retrieval}

To promote higher-quality captions \cite{Kong24}, we rank and filter them using the CLAP similarity~\cite{clap2023}, computed as
\[
\cos(\mathbf{v}_{\text{text}},\, \mathbf{v}_{\text{audio}}),
\]
where $\mathbf{v}_{\text{text}}$ and $\mathbf{v}_{\text{audio}}$ are the CLAP text and audio embeddings, respectively. For each audio clip, we retain the Top-5 captions with the highest similarity scores and discard captions with cosine similarity below 35\% \cite{Kong24}, combining the remaining captions into a single extended caption.


We then leverage these generated captions to perform text-based retrieval—including lexical search, semantic search, and BM25 \cite{bm25}—to identify the closest matches in the text database. 
 The lexical search procedure is as follows:
\begin{enumerate}
    \item Remove stop words from the query and captions.
    \item Count how many query words appear in each caption.
    \item Rank captions by this count, similar to 1-gram BLEU \cite{bleu}.
\end{enumerate}
For semantic search, queries and generated captions were embedded using RoBERTa, and retrieval was based on cosine similarity between query and caption embeddings. 

Retrieval accuracy is measured for both audio-to-text and text-to-audio tasks, with results summarized in Table~\ref{tab:resultsUsingCaption} (due to space constraints, we report only the results for Salmonn, Pengi, GAMA, LTU, Gemini 2.5 Pro, and AudioGPT on the AudioCaps dataset; the complete results will be made available on our GitHub). Compared to our proposed model, this two-step pipeline of caption generation followed by text retrieval exhibits substantially lower performance.



\begin{table*}[ht]
\centering
\resizebox{\textwidth}{!}{%
\begin{tabular}{|l|l|c|c|c|c|c|c|c|c|}
\hline
\textbf{Model} & \textbf{Retrieval} & \textbf{a2t\_R@1} & \textbf{a2t\_R@5} & \textbf{a2t\_R@10} & \textbf{a2t\_mAP@10} & \textbf{t2a\_R@1} & \textbf{t2a\_R@5} & \textbf{t2a\_R@10} & \textbf{t2a\_mAP@10} \\ 
\hline
Salmonn &  Lexical  & 0.125 & 0.201 & 0.220 & 0.138 & 0.176 & 0.294 & 0.392 & 0.147 \\ 
Pengi &  Lexical & 0.059 & 0.147 & 0.204 & 0.037 & 0.128 & 0.307 & 0.362 & 0.184 \\  
GAMA &  Lexical & 0.073 & 0.153 & 0.206 & 0.072 & 0.105 & 0.314 & 0.338 & 0.158 \\  
LTU &  Lexical & 0.092 & 0.173 & 0.288 & 0.084 & 0.172 & 0.303 & 0.422 & 0.201 \\ 
Gemini &  Lexical & 0.202 & 0.289 & 0.388 & 0.153 & 0.193 & 0.393 & 0.428 & 0.241 \\  
AudioGPT& Lexical & 0.183 & 0.295 & 0.572 & 0.093 & 0.165 & 0.387 & 0.392 & 0.219 \\  
Salmonn &  Semantic  & 0.172 & 0.216 & 0.303 & 0.198 & 0.179 & 0.386 & 0.472 & 0.200 \\ 
Pengi &  Semantic & 0.081 & 0.186 & 0.275 & 0.095 & 0.173 & 0.352 & 0.399 & 0.186 \\  
GAMA &  Semantic & 0.093 & 0.184 & 0.238 & 0.093 & 0.128 & 0.393 & 0.408 & 0.199 \\  
LTU &  Semantic & 0.101 & 0.289 & 0.372 & 0.097 & 0.189 & 0.395 & 0.467 & 0.218 \\ 
Gemini &  Semantic & 0.214 & 0.385 & 0.428 & 0.172 & 0.218 & 0.481 & 0.504 & 0.317 \\  
AudioGPT& Semantic & 0.216 & 0.373 & 0.623 & 0.104 & 0.184 & 0.472 & 0.485 & 0.288 \\  

Salmonn &  BM25  & 0.195 & 0.287 & 0.382 & 0.209 & 0.195 & 0.481 & 0.596 & 0.228 \\ 
Pengi &  BM25 & 0.100 & 0.207 & 0.342 & 0.102 & 0.209 & 0.503 & 0.532 & 0.239 \\  
GAMA &  BM25 & 0.112 & 0.292 & 0.394 & 0.104 & 0.208 & 0.520 & 0.597 & 0.294 \\  
LTU &  BM25 & 0.117 & 0.315 & 0.403 & 0.118 & 0.211 & 0.573 & 0.603 & 0.308 \\ 
Gemini &  BM25 & 0.266 & 0.485 & 0.762 & 0.136 & 0.256 & 0.601 & 0.730 & 0.402 \\  
AudioGPT& BM25 & 0.287 & 0.441 & 0.751 & 0.143 & 0.255 & 0.609 & 0.727 & 0.417 \\  
Proposed &    & \textbf{0.451} & \textbf{0.793} & \textbf{0.905} & \textbf{0.486} & \textbf{0.352} & \textbf{0.715} & \textbf{0.844} & \textbf{0.521} \\
\hline
\end{tabular}
}
\caption{Evaluation results of different models and retrieval methods on the AudioCaps dataset using caption generation followed by text based search.}
\label{tab:resultsUsingCaption}
\end{table*}
 
\subsection{Training Details}
\label{app:trainingdetails}

We train all models using the Adam optimizer, with the learning rate selected via Optuna. Batch sizes range from 4 to 128, depending on dataset size and GPU memory constraints. All experiments are conducted on 8 NVIDIA A100 GPUs with 80GB memory each. The number of training epochs is dataset-dependent, ranging from 2 to 45, with early stopping based on validation performance. Optuna is applied independently to each dataset to tune hyperparameters, including learning rate, batch size, number of projection layers, and loss weights, using validation retrieval performance as the optimization objective.

\end{document}